\documentclass[letterpaper]{article} 
\usepackage{aaai23}  
\usepackage{times}  
\usepackage{helvet}  
\usepackage{courier}  
\usepackage[hyphens]{url}  
\usepackage{graphicx} 
\usepackage{natbib}  
\usepackage{caption} 
\usepackage{amsmath,amsfonts,bm}
\makeatletter
\let\proof\@undefined
\let\endproof\@undefined
\makeatother
\usepackage{amsthm}
\theoremstyle{definition}
\theoremstyle{definition}
\theoremstyle{definition}
\theoremstyle{definition}\newtheorem{theorem}{Theorem}
\theoremstyle{definition}\newtheorem{lemma}{Lemma}
\DeclareMathOperator*{\argmin}{arg\,min}

\makeatletter
\newif\if@restonecol
\makeatother

\usepackage[ruled,vlined,linesnumbered,nofillcomment]{algorithm2e}
\SetKw{KwWhere}{where} 
\SetKwInOut{KwFunc}{func:}
\SetKwInOut{Input}{input}
\SetKwInOut{Output}{output}
\SetKwInOut{Requires}{requires}
\usepackage{multirow}
\usepackage{graphicx}
\usepackage{subfigure}
\usepackage{enumitem}
\usepackage{url}
\def\extraspacing{\vspace{0.0mm} \noindent}

\title{Bespoke: A Block-Level Neural Network Optimization Framework \\for Low-Cost Deployment}
\author {
	Jong-Ryul Lee\textsuperscript{\rm 1},
	Yong-Hyuk Moon\textsuperscript{\rm 1,2}\thanks{Corresponding author.}
}
\affiliations {
	\textsuperscript{\rm 1} Electronics and Telecommunications Research Institute (ETRI), Daejeon, Korea\\
	\textsuperscript{\rm 2} University of Science and Technology (UST), Daejeon, Korea \\
	\{jongryul.lee,yhmoon\}@etri.re.kr
}

\usepackage{bibentry}

\begin{document}

\maketitle

\begin{abstract}
	As deep learning models become popular, there is a lot of need for deploying them to diverse device environments. Because it is costly to develop and optimize a neural network for every single environment, there is a line of research to search neural networks for multiple target environments efficiently. However, existing works for such a situation still suffer from requiring many GPUs and expensive costs. Motivated by this, we propose a novel neural network optimization framework named Bespoke for low-cost deployment. Our framework searches for a lightweight model by replacing parts of an original model with randomly selected alternatives, each of which comes from a pretrained neural network or the original model. In the practical sense, Bespoke has two significant merits. One is that it requires near zero cost for designing the search space of neural networks. The other merit is that it exploits the sub-networks of public pretrained neural networks, so the total cost is minimal compared to the existing works. We conduct experiments exploring Bespoke's the merits, and the results show that it finds efficient models for multiple targets with meager cost.
\end{abstract}

\section{Introduction}
Due to the great success of deep learning, neural networks are deployed into various environments such as smartphones, edge devices, and so forth.
In addition, lots of techniques have been proposed for optimizing neural networks for fast inference and small memory usage \cite{park20,liu21,luo20,yu21,lee20,peng19,Hinton2015,Yim17}.

One of the major issues about optimizing neural networks for efficient deployment is that it is not cheap to search for an optimized model for even a single target environment.
If the target environment is changed, we need to search for another optimized model again.
That is, if we have 100 target environments, we have to do optimization 100 times separately.
This situation is not preferred in practice due to the expensive model search/training cost.

There is a line of research for addressing this issue.
Cai et al. \shortcite{cai20} proposed an efficient method, called Once-For-All(OFA), producing non-trivial neural networks simultaneously for multiple target environments.
Sahni et al. \shortcite{sahni21} proposed a method called CompOFA which is more efficient than OFA in terms of training cost and search cost based on reducing the design space for model configuration.
Since these methods do not utilize a pretrained network, they require too much time for training the full network.
In order to address such an expensive cost, Molchanov et al. \shortcite{molchanov22} proposed a latency-aware network acceleration method called LANA based on a pretrained teacher model.
This method dramatically reduces training and search costs with large design space.
In LANA, for each layer in the teacher model, more than 100 alternative blocks (sub-networks) are trained to mimic it.
After training them, LANA replaces ineffective layers in the teacher model with lightweight alternatives using integer programming.
Even if LANA utilizes a pretrained teacher model, it requires designing alternative blocks.
In addition, training many alternatives for every single layer may increase peak CPU/GPU memory usage.
This limitation leads LANA to require $O(M)$ epochs for training the alternatives on ImageNet where $M$ is the number of alternatives for each layer.

\begin{figure*}[t]
	\centering
	\subfigure[Once-For-All (OFA) \cite{cai20}]{
		\includegraphics[scale =0.54] {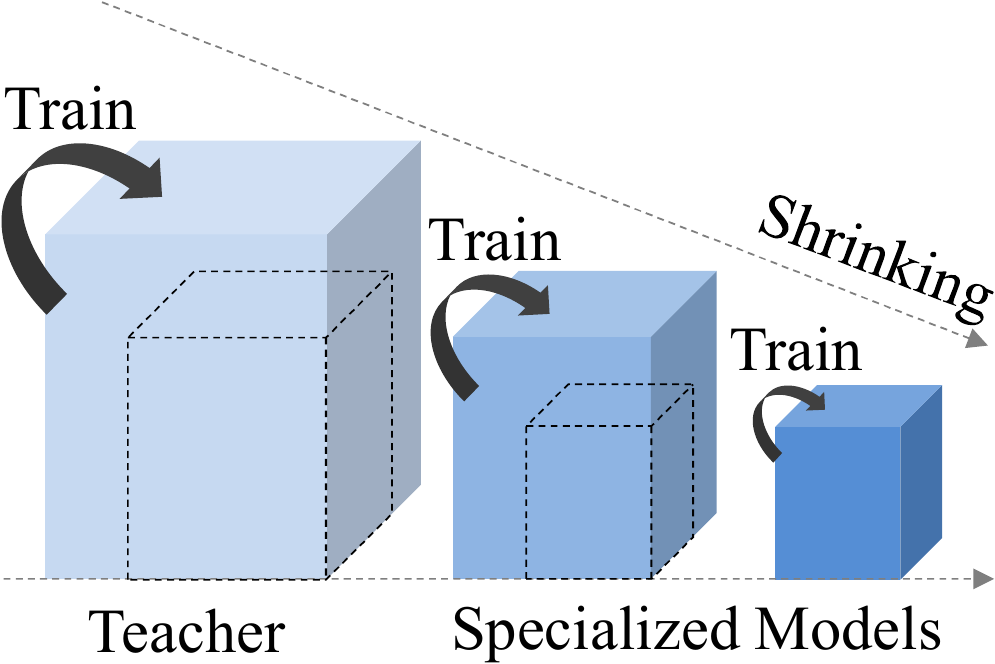}
		\label{fig:pc1u}
	}
	\quad
	\subfigure[LANA \cite{molchanov22}]{
		\includegraphics[scale =0.54] {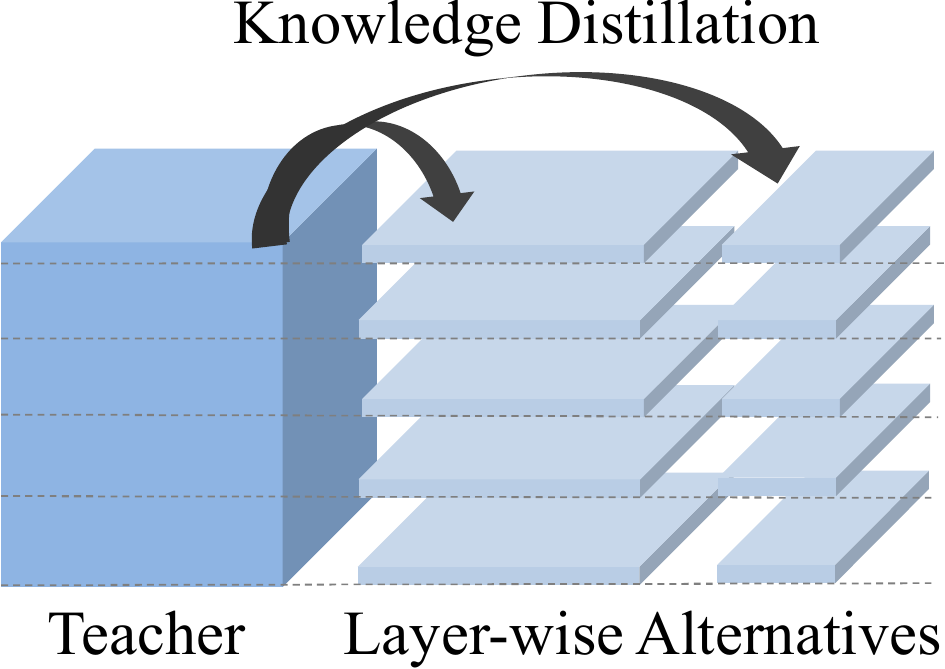}
		\label{fig:pc2u}
	}
	\quad
	\subfigure[Bespoke (This work)]{
		\includegraphics[scale =0.54] {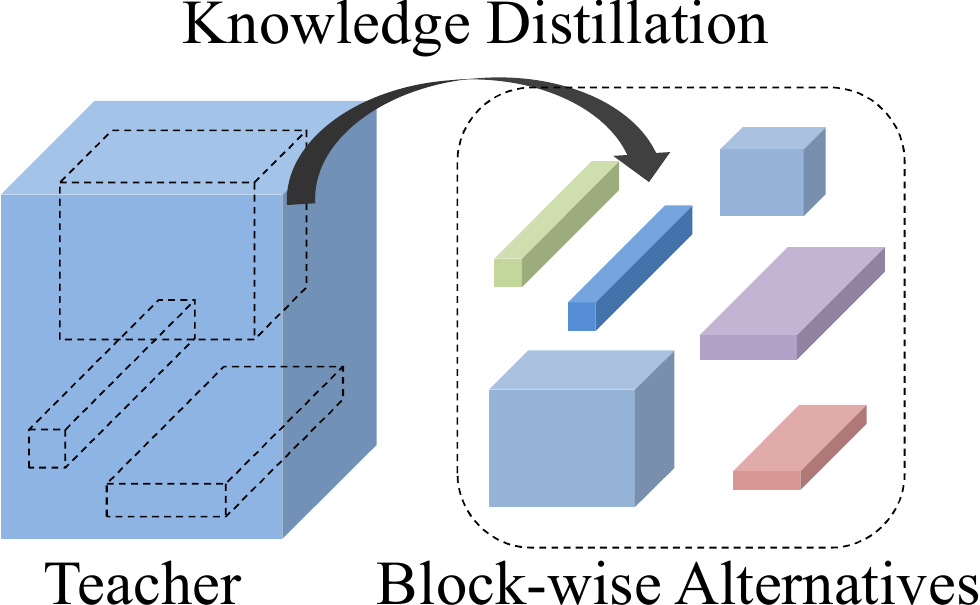}
		\label{fig:pc3u}
	}
	\caption{Illustration of model optimization frameworks for multiple deployment targets}
	\label{fig:frameworks}
\end{figure*}

To significantly reduce the cost of searching for and retraining models for multiple target environments,
we focus on public pretrained neural networks like models in Keras' applications\footnote{https://keras.io/api/applications}.
Each of those networks was properly searched and trained with plentiful resources.
Thus, instead of newly exploring large search space with a number of expensive GPUs, we devise a novel way of exploiting decent reusable blocks (sub-networks) in those networks.
Our work starts from this rationale.

In this paper, we devise a framework named Bespoke efficiently optimizing neural networks for various target environments.
This framework is supposed to help us to get fast and small neural networks with practically affordable cost, which a single modern GPU can handle.
The illustration of comparing OFA, LANA, and Bespoke is depicted in Figure~\ref{fig:frameworks}.
Our framework has following notable features compared to previous methods like OFA and LANA.

\begin{itemize}
	\item \textbf{Near Zero-Cost Model Design Space}: Our framework is based on the sub-networks of a teacher (original) model and those of pretrained models.
	Our framework extracts uniformly at random sub-networks from the teacher and the pretrained networks, and uses them for constructing a lightweight model.	
	Thus, we do not need to struggle to expand the design space with fancy blocks and numerous channel/spatial settings.
	Such blocks are already included in the pretrained neural networks which were well-trained with large-scale data.
	We can use them for a target task with a minor cost of knowledge distillation.
	To the best of our knowledge, this is the first work to utilize parts of such pretrained networks for student model construction.
	
	\item \textbf{Low-Cost Preprocessing}: 
	The randomly selected sub-networks are trained to mimic a part of the teacher via knowledge distillation in preprocessing.
	They can be simultaneously trained with low peak memory usage compared to OFA, CompOFA, and LANA.
	
	\item \textbf{Efficient Model Search}: Our framework deals with a neural network as a set of sub-networks.
	Each sub-network is extracted from the original network so that we can precisely measure the actual inference time of it.
	In addition, for evaluating the suitability of an alternative sub-network,	
	our framework produces an induced neural network by rerouting the original network to use the alternative.
	Then, we compute the accuracy of the induced network for a validation dataset.
	The measured inference time and the accuracy lead us to make an incremental algorithm to evaluate a candidate neural network.
	Based on this, we devise an efficient model search method.
\end{itemize}


This article is the extended version of our AAAI paper\footnote{https://ojs.aaai.org/index.php/AAAI/article/view/26020}.

\section{Related Work}
\extraspacing{\bf Block-wise NAS.}
As our framework deals with a neural network in a block-wise way, there are several early works for block-wise architecture search/generation \cite{zhong18, li20}.
Such works find network architectures at block-level to effectively reduce search space.
Zhong et al. \shortcite{zhong18} proposed a block-wise network generation method with Q-learning, but it is still somewhat slow because it requires training a sampled model from scratch for evaluation.
Li et al. \shortcite{li20} proposed another block-wise method to ensure that potential candidate networks are fully trained without additional training.
Since the potential candidate networks are fully trained, the time for evaluating each architecture can be significantly reduced.
For this, Li et al. \shortcite{li20} used a teacher model and knowledge distillation with it.
Compared to \cite{li20}, we do not only use a teacher model for knowledge distillation but also use it as a base model to find a result model.
In addition, while Li et al. \shortcite{li20} only considered sequentially connected blocks, this work does not have such a limitation regarding the form of a block.
Since we reuse the components (blocks) in the teacher model with their weights, our framework can have lower costs for pretraining and searching than such NAS-based approaches, even if they are block-wisely formulated.

\extraspacing{\bf Hardware-aware NAS.}
Hardware-aware architecture search has been one of the popular topics in deep learning, so there are many existing works \cite{cai20, sahni21, yu20, Moons21, abdelfattah21, molchanov22}.
Those works usually aim to reduce model search space, predict the accuracy of a candidate model, or consider actual latency in the optimization process.
In addition, some of such works were proposed to reduce the cost of deploying models to multiple hardware environments \cite{cai20, sahni21, Moons21, molchanov22}.
Moons et al. \shortcite{Moons21} proposed a method named DONNA built on block-wise knowledge distillation for rapid architecture search with consideration of multiple hardware platforms.
DONNA uses teacher blocks defined in a pre-trained reference model for distilling knowledge to student blocks, not for building student block architectures.
Despite such many existing works, we think that there is an opportunity for improvement in terms of searching/training costs with the sub-networks in pretrained models.

\extraspacing{\bf Knowledge Distillation.}
Our framework includes knowledge distillation to train the sub-networks. 
Some early works for block-wise knowledge distillation which are related to our framework \cite{Yim17, wang18, li20distill, Shen21}.
Shen et al. \shortcite{Shen21} proposed a block-wise distilling method for a student model which is given by model compression or an initialized lightweight model.
In contrast to \cite{li20, Moons21}, they used a teacher model for making a student model.
However, while our framework deals with any sub-network for a single input and a single output as a block,
they used sequentially connected blocks at a high level.
That is, student models produced by our framework are likely to be more diverse than those by \cite{Shen21}.

\section{Method}
\begin{figure}[t]
	\begin{center}
		\includegraphics[scale=0.63]{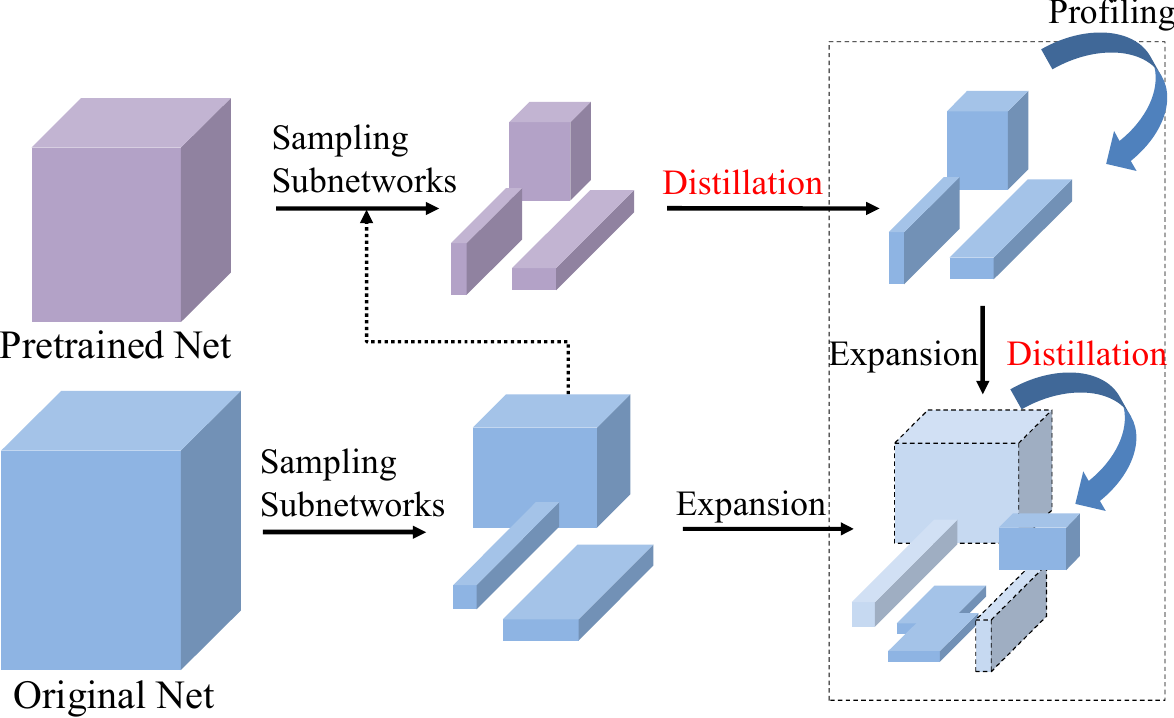}
	\end{center}
	\caption{The flow of the preprocessing step}
	\label{fig:preprocessing}
\end{figure}

Bespoke consists of two steps: preprocessing and model search.
Given a teacher (original) model, the preprocessing step finds sub-networks which will be used as alternatives.
This step is described in Figure~\ref{fig:preprocessing}.
The model search step constructs a student model by replacing some parts of the teacher model with the sub-networks.
In this section, all the proofs are included in the appendix.

\subsection{Formulation}
For a neural network denoted by $\mathcal{N}$, the set of all layers in $\mathcal{N}$ is denoted by $L(\mathcal{N})$.
Then, for any subset $X\subseteq L(\mathcal{N})$, we define a sub-network of $\mathcal{N}$ which consists of layers in $X$ with the same connections in $\mathcal{N}$.
Its inputs are defined over the connections from layers in $L(\mathcal{N})\setminus X$ to $X$ in $\mathcal{N}$.
The outputs of the sub-network are similarly defined.
For simplicity, we consider only sub-networks having a single input and a single output.
We denote a set of sub-networks in $\mathcal{N}$ sampled in a certain way by $\Omega({\mathcal{N}})$.

For a sub-network $\mathcal{S}$, the spatial change of $\mathcal{S}$ is defined to be the ratio between its spatial input size and its spatial output size.
For example, if $\mathcal{S}$ has a convolution with stride 2, its spatial change is $\frac{1}{2}$.
Then, for any two sub-networks $\mathcal{S}$ and $\mathcal{A}$, 
we define that if the spatial change of $\mathcal{S}$ is equal to that of $\mathcal{A}$ and the in/out channels of $\mathcal{A}$ are larger than or equal to those of $\mathcal{S}$, $\mathcal{A}$ is compatible with $\mathcal{S}$.

Suppose that we have a neural network $\mathcal{N}$ and a sub-network $\mathcal{S}$ of $\mathcal{N}$.
Suppose that we have another sub-network $\mathcal{A}$ that is compatible with $\mathcal{S}$ and replace $\mathcal{S}$ with $\mathcal{A}$ for $\mathcal{N}$.
$\mathcal{A}$ may come from $\mathcal{N}$ or another neural network.
If the computational cost of $\mathcal{A}$ is smaller than that for $\mathcal{S}$, such a replacement can be seen as model acceleration/compression.
One can worry a case that the input/output channel dimensions of $\mathcal{S}$ are smaller than those of $\mathcal{A}$.
Even in this case, by pruning input/output channels of $\mathcal{A}$, we can make a pruned network $\mathcal{A}'$ from $\mathcal{A}$ having the same input/output channel dimensions as $\mathcal{S}$.
Then, we can replace $\mathcal{S}$ with $\mathcal{A}'$ in $\mathcal{N}$ without computational violation.
We say that a neural network produced by such a replacement is induced from $\mathcal{N}$.

Suppose that we have a teacher neural network $\mathcal{T}$ which we want to optimize.
Let us consider a set of public pretrained networks, denoted by $\mathbf{P}$.
Then, suppose that we have sub-networks in $\mathcal{T}$ as well as sub-networks in the pretrained networks of $\mathbf{P}$.
By manipulating such sub-networks, we propose a novel structure named a model house $\mathsf{H}=(\Omega(\mathcal{T}), \mathbf{A})$ where $\mathbf{A}$ is a set of sub-networks each of which is derived from $\mathcal{T}$ or a neural network in $\mathbf{P}$.
$\mathbf{A}$ is called the alternative set, and every element in it is compatible with a sub-network in $\Omega(\mathcal{T}$).
Given a model house $\mathsf{H}$, Our framework finds a lightweight model $\mathcal{M^{*}}$ which is induced from $\mathcal{T}$ as follows:
\begin{align}
\mathcal{M^{*}} = \argmin_{\mathcal{M}\in \textit{satisfy}(\mathsf{H}, R)} Loss_{val}(\mathcal{M}),
\end{align}
where $Loss_{val}(\mathcal{M})$ is the task loss of a (neural network) model $\mathcal{M}$ upon the validation dataset, $R$ represents a certain requirement, and $\textit{satisfy}(\mathsf{H}, R)$ is a set of candidate neural networks induced from $\mathcal{T}$ with $\mathsf{H}$ which satisfy $R$.

\begin{figure*}[t]
	\centering
	\subfigure[Simple skip connection]{
		\includegraphics[scale =0.65] {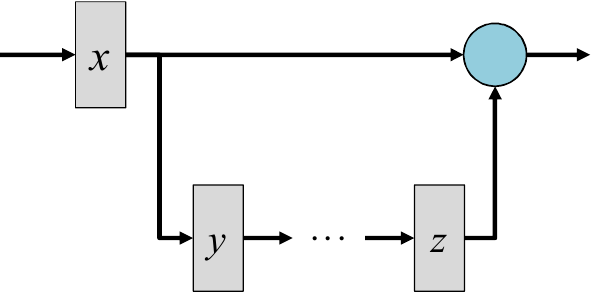}
		\label{fig:residual}
	}
	\qquad
	\subfigure[Nested skip connection]{
		\includegraphics[scale =0.65] {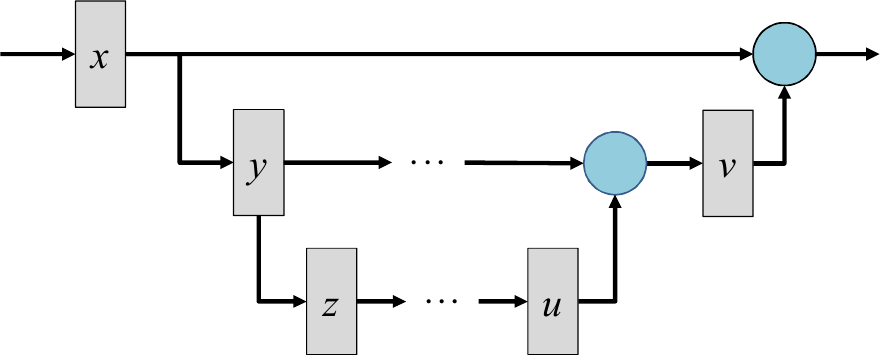}
		\label{fig:deeperconn}
	}
	\qquad
	\subfigure[Sequentially connected blocks]{
		\includegraphics[scale =0.65] {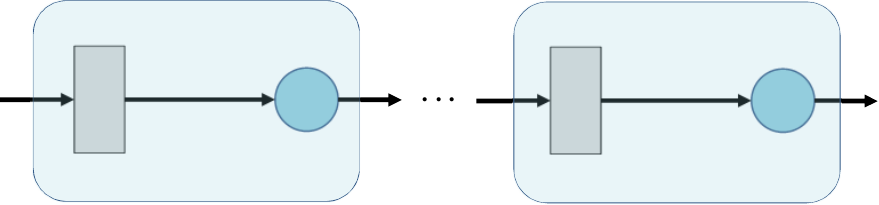}
		\label{fig:blocks}
	}
	\caption{Illustration of example sub-networks having a single input and a single output with skip connections. A rectangle represents a layer and a circle represents an element-wise operation like addition and multiplication.}
	\label{fig:subnets}
\end{figure*}

\subsection{Preprocessing}
Let us introduce how to find effective sub-networks for a model house $\mathsf{H}$ and how to process them.

\extraspacing{\bf Enumeration.}
Let us introduce a way of finding a sub-network in a (teacher) neural network $\mathcal{T}$.
Note that it is not trivial to find such a sub-network in $\mathcal{T}$ due to complex connections between layers such as skip connections.
To address this issue, we use a modified version of the Depth-First Search (DFS).
The modified algorithm starts from any layer in $\mathcal{T}$, and it can move from a current layer $l$ to its outbound neighbors only if all the inbound layers of $l$ were already visited.
For understanding, we provide the detailed procedure of the modified DFS in Algorithm~\ref{algo:ddfs}.
In this algorithm, $|\mathcal{T}|$ represents the number of layers in $\mathcal{T}$.
In addition, for any layer $l$, $\mathbf{N}_{\text{out}}(l)$ is the outbound layers of $l$,
and $\mathbf{N}_{\text{in}}(l)$ is the inbound layers of $l$.
By maintaining a vector $\mathbb{\delta}$, we can make $u$ wait until all the inbound layers of $u$ are visited.

\begin{algorithm}[t]
	\caption{Modified Depth-First Search($l_{0}$, $\mathcal{T}$)}
	\label{algo:ddfs}
	\KwIn{$l_{0}$: a layer in $\mathcal{T}$, $\mathcal{T}$: a neural network}
	\Begin{
		Initialize a stack $S$ with $l_{0}$\;
		Initialize a vector $\mathbb{\delta}$ as the zero vector in $\mathbb{R}^{|\mathcal{T}|}$\;
		\While{$S$ is not empty}
		{
			$v:= \text{pop}(S)$\;
			mark $v$ as being visited\;
			\For{$u \in \mathbf{N}_{\text{out}}(v)$} {						
				Increase $\mathbb{\delta}[u]$ by 1\;
				\If {$\mathbb{\delta}[u] = |\mathbf{N}_{\text{in}}(u)|$} {
					Push $u$ into $S$\;	
				}
			}
		}
	}
\end{algorithm}

Suppose that the modified DFS algorithm started from a layer $l_{0}$.
Let us denote the $(i+1)$-th layer popped from $S$ by $l_{i}$
and the set of popped layers before $l_{i}$ by $P_{i}$.
Then, the modified DFS algorithm guarantees the following lemma.

\begin{lemma} \em
	\label{lemma:single}
	Suppose that $l_{0}$ has a single input and $l_{i}$ has a single output.
	If $l_{i}$ is the only element in the stack when $l_{i}$ is being popped from the traversing stack,
	the sub-network defined by $P_{i}\cup \{l_{i}\}$ from $\mathcal{T}$ whose $l_{i}$ is the output layer has the single input and the single output.
\end{lemma}

The key point of this lemma is the completeness that every layer in $P_{i}\cup \{l_{i}\}$ is connected from or to another in the set except $l_{0}$ and $l_{i}$.
Based on this lemma, we can derive the following theorem.

\begin{theorem} \em
	\label{theorem:all}
	All the possible sub-networks of a neural network $\mathcal{T}$ having a single input and a single output can be enumerated by running the modified DFS algorithm multiple times.
\end{theorem}

For understanding, Figure~\ref{fig:subnets} depicts sub-networks that can be found with the modified DFS algorithm.
An example of the simple skip connection is a sub-network with residual connections in ResNet \cite{he16eccv}.
An example of the nested skip connections is a sub-network with a residual connection and a squeeze-and-excitation module in EfficientNet \cite{tan19}.
Our enumeration algorithm can also find a sub-network including multiple blocks with skip connections depicted in Figure~\ref{fig:blocks}.


\extraspacing{\bf Sampling.}
Because a neural network usually has more than 100 layers, enumerating all possible sub-networks is simply prohibitive with respect to time and space.
Another decent option is to randomly sample sub-networks.
Given a neural network $\mathcal{N}$, the sub-network random sampling algorithm works as follows.
First, it collects all individual layers in $\mathcal{N}$.
Then, it uniformly at random selects a layer among them and conducts the modified DFS algorithm.
Whenever $l_{i}$ is being popped from the stack only having $l_{i}$, the algorithm stores a pair of $l_{i}$ and $P_{i}$.
After the traversal, our algorithm uniformly at random selects one among the stored pairs and derives a sub-network as the return.
In practice, for this random selection, we use a part of the stored pairs close to $l_{0}$ to avoid selecting over-sized sub-networks.
The size ratio of the part to the entire stored pairs is parameterized to $r$.
We denote the sub-network sampling algorithm with a neural network $\mathcal{N}$ by \textit{SubnetSampling}($\mathcal{N}$).

\begin{algorithm}[t]
	\caption{\textit{Construct}($\mathcal{T}$, $\mathbf{P}$, $n_{\mathcal{T}}$, $n_{\mathbf{P}}$)}
	\label{algo:sampling}
	\KwIn{$\mathcal{T}$: the teacher network, $\mathbf{P}$: the set of pretrained networks, $n_{\mathcal{T}}$: the number of sampled sub-networks in $\mathcal{T}$,  $n_{\mathbf{P}}$: the number of sampled sub-networks over $\mathbf{P}$}
	\KwOut{$\Omega(\mathcal{T})$: a set of sampled sub-networks in $\mathcal{T}$, $\Omega(\mathbf{P})$: a set of sampled sub-networks in the neural networks of $\mathbf{P}$, $\textit{TMap}$: a map}
	\Begin{
		
		$\Omega_{\mathcal{T}}$ := $\{\}$, $\Omega_{\mathbf{P}}$ := $\{\}$, initialize $\textit{TMap}$\;
		\While{$|\Omega_{\mathcal{T}}| <  n_{\mathcal{T}}$}
		{
			$\mathcal{S}$ := \textit{SubnetSampling}($\mathcal{T}$)\;
			Add $\mathcal{S}$ into $\Omega_{\mathcal{T}}$\;
		}
		
		\While{$|\Omega_{\mathbf{P}}| <  n_{\mathbf{P}}$}
		{	
			Randomly sample $\mathcal{S} \in \Omega_{\mathcal{T}}$\;
			Randomly sample $\mathcal{P} \in \mathbf{P}$\;
			$\mathcal{P'}$ := \textit{SubnetSampling}($\mathcal{P}$)\;
			\If{$\mathcal{P'}$ is compatible with $\mathcal{S}$}{
				Add $\mathcal{P'}$ into $\Omega_{\mathbf{P}}$\;
				$\textit{TMap}[\mathcal{P'}]:=\mathcal{S}$\;
			}
		}
		\Return $\Omega_{\mathcal{T}}$, $\Omega_{\mathbf{P}}$, $\textit{TMap}$\;
	}
\end{algorithm}

The sub-network sampling algorithm is used to get the sub-networks of a teacher network and pretrained networks.
Let us denote a set of sampled sub-networks of pretrained networks in $\mathbf{P}$ by $\Omega(\mathbf{P})$.
Algorithm~\ref{algo:sampling} describes the overall procedure to find $\Omega(\mathcal{T})$ and $\Omega(\mathbf{P})$ with \textit{SubnetSampling}$(\cdot)$.
In Lines~3-5, it computes sub-networks in $\mathcal{T}$.
In Lines~6-12, for a randomly sampled sub-network $\mathcal{S}\in \Omega_{\mathcal{T}}$, it uniformly at random samples a pretrained neural network $\mathcal{P}$ and gets a sub-network $\mathcal{P'}$ of it.
We store a mapping from $\mathcal{P'}$ to $\mathcal{S}$ into $\textit{TMap}$ for our search algorithm.


\extraspacing{\bf Further Approximation.}
In order to expand the search space of our framework, we generate more sub-networks and add them to $\mathbf{A}$.
One straightforward approach is to apply model compression techniques such as channel pruning and decomposition to the sub-networks in $\Omega(\mathcal{T})$ and $\Omega(\mathbf{P})$.
For this purpose, we use a variant of a discrepancy-aware channel pruning method \cite{luo20} named CURL.
This variant simply uses mean-squared-error as the objective function, while the original version uses KL divergence.

Since a channel pruning method changes the number of the in/out channels of sub-networks,
one can ask about how to handle the compatibility between sub-networks.
While every alternative should have the same in/out shapes of a teacher layer in \cite{molchanov22},
we permit the changes of an alternative's in/out shapes by introducing channel-wise masking layers.

To implement this, for any alternative sub-network $\mathcal{A}\in \mathbf{A}$, we append an input masking layer $M_{\text{in}}$ and output masking layer $M_{\text{out}}$ after the input layer and the output layer, respectively.
If a channel pruning method makes channel-wise masking information for the in/out channels of $\mathcal{A}$, it is stored in $M_{\text{in}}$ and $M_{\text{out}}$.
The masked version of $\mathcal{A}$ is used instead of $\mathcal{A}$ when $\mathcal{A}$ is selected as a replacement.
Suppose that a resulting student model has such masked sub-networks.
After fine-tuning the student model, we prune internal channels in it according to the masking information in masking layers.
Keeping the masking layers during fine-tuning is essential for intermediate feature-level knowledge distillation.
For understanding, we provide figures explaining this procedure in the supplemental material.

We denote a set of networks added by the expansion process from $\Omega(\mathcal{T})$ and $\Omega(\mathbf{P})$ by $\Omega^{*}(\mathcal{T}, \mathbf{P})$.
Then, the alternative set $\mathbf{A}$ is defined as,
\begin{align}
\mathbf{A} = \Omega(\mathcal{T}) \cup \Omega(\mathbf{P}) \cup \Omega^{*}(\mathcal{T}, \mathbf{P}).
\end{align}

\extraspacing{\bf Sub-Network Training and Profiling.}
Let us discuss how to train the sub-networks in $\mathbf{A}$ to use them for producing a student model.
We train them via knowledge distillation as Molchanov et al. \shortcite{molchanov22} did.
Recall that each alternative sub-network $\mathcal{A}\in \mathbf{A}$ has a corresponding sub-network $\mathcal{S}\in \Omega(\mathcal{T})$.
Thus, for each sub-network $\mathcal{A}\in \mathbf{A}$, we can construct a distillation loss defined by the mean-squared-error between the output of $\mathcal{A}$ and that of $\mathcal{S}$.
By minimizing such a MSE loss by gradient descent optimization, we can train the sub-networks in $\mathbf{A}$.
Formally, the mean-square-error loss for training alternatives is defined as follows:
\begin{align}
\min_{W_{\mathbf{A}}} \sum_{x\in X_{\text{sample}}} \sum_{\mathcal{S}\in \Omega(\mathcal{T})} \sum_{\mathcal{A}\in \mathbf{A}_{\mathcal{S}}} \left(f_{\mathcal{S}}(x^{*}) - f_{\mathcal{A}}(x^{*})\right)^{2},
\end{align}
where $W_{\mathbf{A}}$ is the set of all weight parameters over $\mathbf{A}$, $X_{\text{sample}}$ is a set of sampled input data, $\mathbf{A}_{\mathcal{S}}$ is a set of alternatives compatible with $\mathcal{S}$, $f_{\mathcal{S}}$ is the functional form of $\mathcal{S}$, and $x^{*}$ is the output of the previous layer of $\mathcal{S}$ in $\mathcal{T}$.
It is noteworthy that if we have enough memory size for a GPU, all the sub-networks in $\mathbf{A}$ can be trained together in a single distillation process.
If GPU memory is insufficient, they can be trained over several distillation processes.

\extraspacing{\bf Requirement and Profiling.}
A requirement can be related to FLOPs, actual latency, memory usage, and so forth.
In order to produce a student model satisfying a requirement, we measure such metrics for every alternative sub-network after the sub-network training step.

\subsection{Model Search}
Given a requirement and a model house $\mathsf{H}=(\Omega(\mathcal{T}), \mathbf{A})$, our search algorithm for a student model is based on simulated annealing.
Simulating annealing is a meta-heuristic algorithm for optimizing a function within computational budget.
In order to exploit simulated annealing, we need to define three things: an initial solution, a function to define a next candidate, and the objective function.
For simulated annealing, a solution $\mathbf{S}$ is defined to be a set of sub-networks in $\mathbf{A}$, and we say that $\mathbf{S}$ is feasible when sub-networks in $\mathcal{T}$ corresponded to the sub-networks in $\mathbf{S}$ by $\textit{TMap}$ are not overlapped.
In this work, the initial solution is a solution selected by a simple greedy method.
For a solution $\mathbf{S}$, the score function for the greedy method is defined as,
\begin{align}
\Delta acc(\mathbf{S}) &= \sum_{\mathcal{A}\in \mathbf{S}} \Delta acc(\mathcal{A}, \textit{TMap}[\mathcal{A}], \mathcal{T}),\\
\textit{score}(\mathbf{S}) &= \max\left\{1.0, \frac{\textit{metric}(\mathcal{N}_{\mathbf{S}})}{R} \right\} + \lambda \Delta acc(\mathbf{S}),
\end{align}
where $\Delta acc(\mathcal{A}, \textit{TMap}[\mathcal{A}], \mathcal{T})$ represents the accuracy loss when $\textit{TMap}[\mathcal{A}]$ is replaced with $\mathcal{A}$ for $\mathcal{T}$.
The values for $\Delta acc(\cdot)$ can be calculated in preprocessing.
$\textit{metric}(\cdot)$ is a metric function to evaluate the inference time (e.g., latency) of a candidate student model, and $\mathcal{N}_{\mathbf{S}}$ is a candidate student model induced by $\mathbf{S}$.
$R$ is an objective value in terms of $\textit{metric}(\cdot)$.
We define $\textit{score}(\cdot)$ handling a single requirement (metric) for simplicity, but it can be easily extended to multiple requirements.
This score function is also used in the annealing process.
Note that since constructing $\mathcal{N}_{\mathbf{S}}$ with alternatives is time-consuming, we take another strategy for computing it.
$\textit{metric}(\mathcal{N}_{\mathbf{S}})$ is approximated incrementally from $\textit{metric}(\mathcal{T})$ by adding $\textit{metric}(\mathcal{A})$ and subtracting $\textit{metric}(\textit{TMap}[\mathcal{A}])$ for each sub-network $\mathcal{A}\in \mathbf{S}$.

The remaining thing to complete the annealing process is to compute the next candidate solution.
Given a current solution $\mathbf{S}$, we first copy it to $\mathbf{S}'$ and simply uniformly at random remove a sub-network from it.
Then, we repeatedly add a randomly selected sub-network in $\mathbf{A}$ into $\mathbf{S}'$ until the selected sub-network violates the feasibility of $\mathbf{S}'$.
After that, $\mathbf{S}'$ becomes a next candidate solution.

\begin{table*}[t]
	\centering
	\small
	\begin{tabular}[c]{l | c | c | c | c | c | c }
		\hline
		{\multirow{2}{*}{Methods}}			&	\multicolumn{3}{c}{CIFAR-100}	& \multicolumn{3}{c}{ImageNet}	 \\
		&	Top-1 Acc. (\%)	&	Latency (ms)		& Params (M)	&	Top-1 Acc. (\%)	&		Latency (ms) & Params (M)	\\
		\hline
		\hline
		EfficientNetB2							&		88.01		&	72.78		&	7.91		&	79.98	&	28.40	&	9.18	\\
		EfficientNetB0							&		86.38		&	41.28		&	4.18		&	77.19	&	18.63	&	5.33	\\
		Bespoke-EB2-Rand						&		79.45		&	26.11		&	6.98		&	71.78	&	16.72	&	9.58	\\
		Bespoke-EB2-Only						&		84.47		&	33.07		&	4.39		&	65.62	&	13.74	&	6.30	\\		
		\bf{Bespoke-EB2}						&		\bf{85.45}		&	\bf{31.82}		&	\bf{7.43}		&	\bf{78.61}	&	\bf{17.55}	&	\bf{9.39}	\\
		\hline
		MobileNetV3-Large						&		84.33		&	24.71		&	3.09		&	75.68	&	13.54	&	5.51	\\		
		MobileNetV3-Small						&		81.98		&	11.98		&	1.00		&	68.21	&	6.99	&	2.55	\\	
		CURL									&		81.49		&	28.65		&	1.89		&	71.21	&	17.20	&	2.37	\\		
		\bf{Bespoke-EB0}						&		\bf{84.86}		&	\bf{22.05}		&	\bf{3.78}		&	\bf{73.93}	&	\bf{9.39}	&	\bf{6.77}	\\
		
		\hline
	\end{tabular}
	\caption{Overall Results (CPU Latency)}
	\centering
	\label{table:cpu}
\end{table*}

\extraspacing{\bf Retraining with Distillation.}
After getting a student model, we retrain it with knowledge distillation.
Consider that the final solution of the simulated annealing process is denoted by $\mathbf{S}^{*}$.
This distillation process works with the mean-square-error loss between the outputs corresponding to the sub-networks of $\mathbf{S}^{*}$ in $\mathcal{N}_{\mathbf{S}^{*}}$ and their corresponding sub-networks in $\mathcal{T}$.

\section{Experiments}

\subsection{Implementation Detail}
\extraspacing{\bf Datasets.}
We use two representative datasets: CIFAR-100 \cite{cifar100} and ImageNet-1K (ImageNet) \cite{ILSVRC15}.
CIFAR-100 has 50K training images and 10K validation images with 100 classes.
ImageNet has 1,281K training images and 50K validation images with 1000 classes.


\extraspacing{\bf Models.}
As original (teacher) models, we use EfficientNet \cite{tan19}.
As pretrained models, we do not only use them, but also ResNet50V2 \cite{he16eccv}, MobileNet \cite{Howard17}, MobileNetV2 \cite{Sandler18}, MobileNetV3 \cite{Howard_2019_ICCV}, Inception-ResNet \cite{Szegedy17}, ResNetRS \cite{bello2021revisiting}, and EfficientNetV2 \cite{Tan21}.
Those models were already pretrained and are available at Keras' applications.

It should be noticed that for CIFAR-100, EfficientNet and MobileNetV3 are transformed for transfer learning to additionally include a global average pooling layer, a dropout layer and a fully-connected layer.
That is why they are slower for CIFAR-100 than for ImageNet.

\extraspacing{\bf Computation Environment.}
Each experiment for CIFAR-100 is conducted with NVIDIA GeForce Titan RTX,
while that for ImageNet is conducted with NVIDIA RTX A5000.

CPU latency and GPU latency are measured with ONNX Runtime\footnote{https://github.com/onnx/onnx}.
For reference, Intel i9-10900X is used for CIFAR-100, while AMD EPYC 7543 32-Core processor is used for ImageNet.

\extraspacing{\bf Implementation.}
We implement all things with TensorFlow Keras because it is easy to implement the sub-network extraction and injection with Keras.
Our training code is based on NVIDIA Deep Learning Examples\footnote{https://github.com/NVIDIA/DeepLearningExamples}.

EfficientNet and MobileNetV3 are fine-tuned for CIFAR-100 with the learning rate is 0.0001 and cosine annealing \cite{loshchilov2017sgdr}.
We used CutMix \cite{Yun_2019_ICCV} and MixUp \cite{zhang2018mixup} together for fine-tuning them, but not for retraining a student model. 

The learning rate for training alternative sub-networks and fine-tuning a student model after the model search step is 0.02.
We used the SGD optimizer for training the alternatives, and the AdaBelief optimizer \cite{Zhuang20} for fine-tuning the student model.
We used the same hyperparameter values for the AdaBelief optimizer depending on datasets as described in \cite{Zhuang20}.
The number of epochs for training the alternatives is one for ImageNet and 20 for CIFAR-100, respectively.
In addition, the number of epochs for fine-tuning the student model is 15 for ImageNet and 30 for CIFAR-100, respectively.

Recall $r$ is the ratio parameter to control the maximum size of sampled sub-networks.
It is set to 20\% for ImageNet and 30\% for CIFAR-100.
$n_{\mathcal{T}}$ is the parameter for $|\Omega(\mathcal{T})|$ and $n_{\mathbf{P}}$ is the parameter for $|\Omega(\mathbf{P})|$.
For all the experiments, $n_{\mathcal{T}}$ and $n_{\mathbf{P}}$ are fixed to 100 and 200, respectively.
In addition, $|\Omega^{*}(\mathcal{T}, \mathbf{P})|$, which is the number of sub-networks added by the expansion process, is also fixed to 200.

\subsection{Results}
Building a model house and searching for a student model are repeated 5 times.
Thus, each reported latency/accuracy for Bespoke is the average over 25 measured values.

\extraspacing{\bf Overall Inference Results.}
CURL \cite{luo20} is a channel pruning method which gives a global score to each channel in a network.
Based on the score information, we can make models with various scales.
Thus, we compare our framework with it in terms of latency and accuracy.

\begin{table}[t]
	\centering
	\small
	\begin{tabular}[c]{l | c | c}
		\hline
		Methods		&	Top-1 Acc. (\%)	&	Latency (ms)		 \\
		\hline
		\hline
		EfficientNetB2						&		88.01		&	5.78	\\
		EfficientNetB0						&		86.38		&	4.06	\\		
		Bespoke-EB2-Rand					&		79.45		&	5.36	\\		
		Bespoke-EB2-Only					&		87.30		&	5.15	\\												
		\bf{Bespoke-EB2}					&		\bf{87.01}		&	\bf{3.08}	\\				
		\hline
		
		MobileNetV3-Large					&		84.33		&		3.04		\\				
		MobileNetV3-Small					&		81.98		&		2.52		\\								
		CURL								&		81.49		&		4.02		\\										
		\bf{Bespoke-EB0}					&		\bf{86.18}		&		\bf{3.19}	\\										
		\hline
		
	\end{tabular}
	\caption{GPU Latency (CIFAR-100)}
	\centering
	\label{table:gpu}
\end{table}

\begin{table}[t]
	\centering
	\small
	\begin{tabular}[c]{l | c | c}
		\hline
		Methods		&	Top-1 Acc. (\%)	&	Latency (ms)		 \\
		\hline
		\hline
		EfficientNetB2						&		79.98			&	3.55		\\
		EfficientNetB0						&		77.19			&	2.29		\\	
		Bespoke-EB2-Rand					&		71.78		&	2.55	\\		
		Bespoke-EB2-Only					&		74.22		&	2.83	\\			
		\bf{Bespoke-EB2}					&		\bf{78.08}		& 	\bf{2.16}	\\								
		\hline
		MobileNetV3-Large					&		75.68			&	1.82		\\				
		MobileNetV3-Small					&		68.21			&	1.29	\\
		CURL								&		71.21			&	1.81		\\										
		\bf{Bespoke-EB0}					&		\bf{72.75}		&	\bf{2.40}		\\
		\hline
		
	\end{tabular}
	\caption{GPU Latency (ImageNet)}
	\centering
	\label{table:gpu_imagenet}
\end{table}

The overall results are presented in Table~\ref{table:cpu}, Table~\ref{table:gpu}, and Table~\ref{table:gpu_imagenet}.
In these tables, the results of applying Bespoke to EfficientNet-B2 (EfficientNet-B0) are presented with Bespoke-EB2 (Bespoke-EB0).
`Rand' represents the results of models built by random selection on the same model house of Bespoke-EB2.
`Only' represents the results of making Bespoke only use the sub-networks of a teacher.
It is noteworthy that the accuracy of Bespoke-EB2 is higher than that of EfficientNet-B0 for ImageNet while they have similar CPU latency.
This means that Bespoke can effectively scale down EfficientNet-B2 in terms of CPU latency.
The accuracy of Bespoke-EB2 for CIFAR-100 is lower than that of EfficientNet-B0, but Bespoke-EB2 is more efficient than EfficientNet-B0.
Compared to MobileNetV3 models, which were proposed for fast inference, Bespoke-EB2 is somewhat slower than them.
However, it can still be more accurate than the MobileNetV3 models.
Bespoke-EB0 is faster than MobileNetV3-Large with slightly less accuracy.
Note that the accuracy gap between MobileNetV3-Large and Bespoke-EB2 is larger than the gap between MobileNetV3-Large and Bespoke-EB0, while the latency gap is not.
In addition, the latency difference between Bespoke-EB0 and MobileNetV3-Small is marginal, but the accuracy gap is significant.
This means that the student models of Bespoke have a better trade-off between accuracy and latency than those of MobileNetV3.

Models found by Bespoke-EB2-Rand have inferior accuracy compared to the other competitors.
This result supports that selecting good alternatives is important for Bespoke, and the search algorithm of Bespoke effectively works. 
In addition, we observed that models found by Bespoke-EB2-Only are less competitive than those by Bespoke-EB2.
From this result, the public pretrained networks are actually helpful for constructing fast and accurate models.

Meanwhile, Bespoke-EB0 is somewhat behind in terms of a trade-off between accuracy and GPU latency, especially for ImageNet.
This may happen with the quality of the sampled sub-networks.
Bespoke-EB0 will produce better results with a more number of sampled sub-networks and a more number of epochs for training them.

Since CURL does not exploit actual latency, its latency gain is insignificant.
Recall that Bespoke utilizes the variant of CURL to expand the alternative set.
This observation implies that the latency gain of Bespoke mainly comes from replacing expensive sub-networks with fast alternatives, not the expansion.
We expect that Bespoke can find faster models with latency-aware compression methods.

\begin{table}[t]
	\centering
	\small
	\begin{tabular}[c]{l | c | c | c}
		\hline
		
		Methods		&	Preprocess	&	Search	&	Retrain 		\\
		\hline
		\hline
		OFA				&	1205			&	40			&	75			\\
		DNA				&	320			&	14			&	450	\\
		DONNA			&	1920+1500		&	1500		&	50		\\		
		LANA			&	197			&	$<$ 1h		&	100	\\		
		\hline
		\bf{Bespoke}	&	$|\mathbf{A}|/G=4$		&	$<$ 0.5h	&	15		\\		
		\hline
		
	\end{tabular}
	\caption{Cost Analysis for ImageNet (Epochs)}
	\centering
	\label{table:cost}
\end{table}

\extraspacing{\bf Cost analysis.}
We analyze the cost for preprocessing, searching, and retraining.
The results for OFA \cite{cai20}, DNA \cite{li20}, DONNA \cite{Moons21}, and LANA \cite{molchanov22} come from \cite{molchanov22}.
Note that `preprocess' includes pretraining process and `search' is the cost for searching for a single target.

The result is described in Table~\ref{table:cost}.
Note that $G$ is the number of sub-networks that can be trained with a single GPU in parallel.
In the experiments, $G$ is 100 for ImageNet.
Since each sub-network is pretrained with one epoch, the entire cost for pretraining the alternatives is $|\mathbf{A}|/G$.
The preprocessing step includes applying the channel pruning method, but its cost is negligible.
Note that the preprocessing cost of LANA is also calculated with parallelism.
From this table, we can see that Bespoke requires significantly less costs than the other methods for all the steps.
Especially, due to the concept of the block-level replacement, the preprocessing cost of Bespoke is an order of magnitude smaller than that of any other method.

\begin{table}[t]
	\centering
	\small
	\begin{tabular}[c]{l | c | c  }
		\hline
		Methods			&	Top-1 Acc. (\%)	&		Speed-up		\\
		\hline
		\hline		
		EfficientNet-B0		&	77.19				&		1.0	\\		
		\hline
		LANA-EB2-0.4 (CPU)		&		78.11			&		1.19		\\		
		LANA-EB2-0.5 (CPU)		&		78.87			&		0.98		\\		
		\bf{Bespoke-EB2}	(CPU)		&		78.61			&		1.06		\\		
		\hline		
		LANA-EB2-0.45 (GPU)		&		79.71			&		1.10	\\		
		\bf{Bespoke-EB2} (GPU)		&		78.08			&		1.06		\\				
		\hline
		
	\end{tabular}
	\caption{Comparison with LANA (ImageNet)}
	\centering
	\label{table:comp_lana}
\end{table}

\extraspacing{\bf Comparison with LANA.}
We provide an analysis to compare Bespoke with LANA \cite{molchanov22}, which is described in Table~\ref{table:comp_lana}.
Because it is not possible to set up the same environment with LANA, we use relative speed-up from EfficientNet-B0.
The model found by LANA is more accurate than that by Bespoke for GPU latency with faster inference latency.
On the other hand, Bespoke-EB2 is not dominated by the models of LANA in terms of accuracy and CPU latency.
This result is quite promising because Bespoke requires much less cost for preprocessing, but there is a situation that Bespoke is comparable to LANA.

\begin{table}[t]
	\centering
	\begin{tabular}[c]{l | c | c}
		\hline
		
		Ratio		&	Top-1 Acc. (\%)	&	CPU Latency (ms)	\\
		\hline
		\hline
		1.0				&	85.45		&	31.82		\\
		0.75			&	86.23		&	37.34			\\		
		0.5				&	86.23		&	44.50 		\\		
		\hline
		
	\end{tabular}
	\caption{Random Partial Alternative Set Test (CIFAR-100, Bespoke-B2)}
	\centering
	\label{table:partial}
\end{table}

\extraspacing{\bf Random sub-alternative set test.}
We conducted experiments evaluating Bespoke with random subsets of alternative set $\mathbf{A}$ (Table~\ref{table:partial}).
The ratio is the size ratio from a random subset to $\mathbf{A}$.
Bespoke finds an inefficient student model with the random subset of half size, which is slower than EfficientNet-B0.
This is because many efficient and effective sub-networks are not included in the subset.
This result supports the importance of a good alternative set.

\section{Conclusions}
This paper proposes an efficient neural network optimization framework named Bespoke, which works with sub-networks in an original network and public pretrained networks.
The model design space of Bespoke is defined by those sub-networks.
Based on this feature, it can make a lightweight model for a target environment at a very low cost.
The results of the experiments support the assumption that such sub-networks can actually be helpful in reducing searching and retraining costs.
Compared to LANA, Bespoke can find a lightweight model having comparable accuracy and CPU latency. Its entire cost is much lower than that required by LANA and the other competitors.

For future work, we consider making Bespoke support other valuable tasks such as pose estimation and image segmentation.
In addition, by adding more recent pretrained neural architectures into Bespoke, we want to see how Bespoke improves with them.

\section{Acknowledgments}
This work was supported by Institute of Information \& communications Technology Planning \& Evaluation (IITP) grant funded by the Korea government(MSIT) (No. 2021-0-00907, Development of Adaptive and Lightweight Edge-Collaborative Analysis Technology for Enabling Proactively Immediate Response and Rapid Learning).

\bigskip

\bibliography{bespoke}

\begin{thebibliography}{36}
\providecommand{\natexlab}[1]{#1}

\bibitem[{Abdelfattah et~al.(2021)Abdelfattah, Mehrotra, Dudziak, and
  Lane}]{abdelfattah21}
Abdelfattah, M.~S.; Mehrotra, A.; Dudziak, {\L}.; and Lane, N.~D. 2021.
\newblock Zero-Cost Proxies for Lightweight {\{}NAS{\}}.
\newblock In \emph{ICLR}.

\bibitem[{Bello et~al.(2021)Bello, Fedus, Du, Cubuk, Srinivas, Lin, Shlens, and
  Zoph}]{bello2021revisiting}
Bello, I.; Fedus, W.; Du, X.; Cubuk, E.~D.; Srinivas, A.; Lin, T.-Y.; Shlens,
  J.; and Zoph, B. 2021.
\newblock Revisiting ResNets: Improved Training and Scaling Strategies.
\newblock In \emph{NeurIPS}.

\bibitem[{Cai et~al.(2020)Cai, Gan, Wang, Zhang, and Han}]{cai20}
Cai, H.; Gan, C.; Wang, T.; Zhang, Z.; and Han, S. 2020.
\newblock Once-for-All: Train One Network and Specialize it for Efficient
  Deployment.
\newblock In \emph{ICLR}.

\bibitem[{He et~al.(2016)He, Zhang, Ren, and Sun}]{he16eccv}
He, K.; Zhang, X.; Ren, S.; and Sun, J. 2016.
\newblock Identity Mappings in Deep Residual Networks.
\newblock In \emph{ECCV}, 630--645.

\bibitem[{Hinton, Vinyals, and Dean(2015)}]{Hinton2015}
Hinton, G.; Vinyals, O.; and Dean, J. 2015.
\newblock Distilling the Knowledge in a Neural Network.
\newblock \emph{arXiv preprint arXiv:1503.02531}.

\bibitem[{Howard et~al.(2019)Howard, Sandler, Chu, Chen, Chen, Tan, Wang, Zhu,
  Pang, Vasudevan, Le, and Adam}]{Howard_2019_ICCV}
Howard, A.; Sandler, M.; Chu, G.; Chen, L.-C.; Chen, B.; Tan, M.; Wang, W.;
  Zhu, Y.; Pang, R.; Vasudevan, V.; Le, Q.~V.; and Adam, H. 2019.
\newblock Searching for MobileNetV3.
\newblock In \emph{ICCV}, 1314--1324.

\bibitem[{Howard et~al.(2017)Howard, Zhu, Chen, Kalenichenko, Wang, Weyand,
  Andreetto, and Adam}]{Howard17}
Howard, A.~G.; Zhu, M.; Chen, B.; Kalenichenko, D.; Wang, W.; Weyand, T.;
  Andreetto, M.; and Adam, H. 2017.
\newblock MobileNets: Efficient Convolutional Neural Networks for Mobile Vision
  Applications.
\newblock \emph{CoRR}, abs/1704.04861.

\bibitem[{Krizhevsky(2009)}]{cifar100}
Krizhevsky, A. 2009.
\newblock Learning Multiple Layers of Features from Tiny Images.
\newblock \emph{Technical Report TR-2009, University of Toronto}.

\bibitem[{Lee et~al.(2020)Lee, Hong, Hong, and Kim}]{lee20}
Lee, C.; Hong, S.; Hong, S.; and Kim, T. 2020.
\newblock Performance analysis of local exit for distributed deep neural
  networks over cloud and edge computing.
\newblock \emph{ETRI Journal}, 42(5): 658--668.

\bibitem[{Li et~al.(2020{\natexlab{a}})Li, Peng, Yuan, Wang, Liang, Lin, and
  Chang}]{li20}
Li, C.; Peng, J.; Yuan, L.; Wang, G.; Liang, X.; Lin, L.; and Chang, X.
  2020{\natexlab{a}}.
\newblock Block-Wisely Supervised Neural Architecture Search With Knowledge
  Distillation.
\newblock In \emph{CVPR}, 1986--1995.

\bibitem[{Li et~al.(2020{\natexlab{b}})Li, Li, Liu, and Zhang}]{li20distill}
Li, T.; Li, J.; Liu, Z.; and Zhang, C. 2020{\natexlab{b}}.
\newblock Few Sample Knowledge Distillation for Efficient Network Compression.
\newblock In \emph{CVPR}, 14627--14635.

\bibitem[{Lin et~al.(2014)Lin, Maire, Belongie, Bourdev, Girshick, Hays,
  Perona, Ramanan, Zitnick, and Dollár}]{lin2014microsoft}
Lin, T.-Y.; Maire, M.; Belongie, S.; Bourdev, L.; Girshick, R.; Hays, J.;
  Perona, P.; Ramanan, D.; Zitnick, C.~L.; and Dollár, P. 2014.
\newblock Microsoft COCO: Common Objects in Context.

\bibitem[{Liu et~al.(2021)Liu, Zhang, Kuang, Zhou, Xue, Wang, Chen, Yang, Liao,
  and Zhang}]{liu21}
Liu, L.; Zhang, S.; Kuang, Z.; Zhou, A.; Xue, J.-H.; Wang, X.; Chen, Y.; Yang,
  W.; Liao, Q.; and Zhang, W. 2021.
\newblock Group Fisher Pruning for Practical Network Compression.
\newblock In \emph{ICML}, volume 139, 7021--7032.

\bibitem[{Loshchilov and Hutter(2017)}]{loshchilov2017sgdr}
Loshchilov, I.; and Hutter, F. 2017.
\newblock {SGDR}: Stochastic Gradient Descent with Warm Restarts.
\newblock In \emph{ICLR}.

\bibitem[{Luo and Wu(2020)}]{luo20}
Luo, J.-H.; and Wu, J. 2020.
\newblock Neural Network Pruning With Residual-Connections and Limited-Data.
\newblock In \emph{CVPR}, 1455--1464.

\bibitem[{Molchanov et~al.(2022)Molchanov, Hall, Yin, Kautz, Fusi, and
  Vahdat}]{molchanov22}
Molchanov, P.; Hall, J.; Yin, H.; Kautz, J.; Fusi, N.; and Vahdat, A. 2022.
\newblock LANA: Latency Aware Network Acceleration.
\newblock In \emph{ECCV}, 137–156.

\bibitem[{Moons et~al.(2021)Moons, Noorzad, Skliar, Mariani, Mehta, Lott, and
  Blankevoort}]{Moons21}
Moons, B.; Noorzad, P.; Skliar, A.; Mariani, G.; Mehta, D.; Lott, C.; and
  Blankevoort, T. 2021.
\newblock Distilling Optimal Neural Networks: Rapid Search in Diverse Spaces.
\newblock In \emph{ICCV}, 12209--12218.

\bibitem[{Park et~al.(2020)Park, Lee, Mo, and Shin}]{park20}
Park, S.; Lee, J.; Mo, S.; and Shin, J. 2020.
\newblock Lookahead: {A} Far-sighted Alternative of Magnitude-based Pruning.
\newblock In \emph{ICLR}.

\bibitem[{Peng et~al.(2019)Peng, Wu, Chen, and Huang}]{peng19}
Peng, H.; Wu, J.; Chen, S.; and Huang, J. 2019.
\newblock Collaborative Channel Pruning for Deep Networks.
\newblock In \emph{ICML}, volume~97, 5113--5122.

\bibitem[{Russakovsky et~al.(2015)Russakovsky, Deng, Su, Krause, Satheesh, Ma,
  Huang, Karpathy, Khosla, Bernstein, Berg, and Fei-Fei}]{ILSVRC15}
Russakovsky, O.; Deng, J.; Su, H.; Krause, J.; Satheesh, S.; Ma, S.; Huang, Z.;
  Karpathy, A.; Khosla, A.; Bernstein, M.; Berg, A.~C.; and Fei-Fei, L. 2015.
\newblock {ImageNet Large Scale Visual Recognition Challenge}.
\newblock \emph{International Journal of Computer Vision (IJCV)}, 115(3):
  211--252.

\bibitem[{Sahni et~al.(2021)Sahni, Varshini, Khare, and Tumanov}]{sahni21}
Sahni, M.; Varshini, S.; Khare, A.; and Tumanov, A. 2021.
\newblock Comp{\{}OFA{\}} {\textendash} Compound Once-For-All Networks for
  Faster Multi-Platform Deployment.
\newblock In \emph{ICLR}.

\bibitem[{Sandler et~al.(2018)Sandler, Howard, Zhu, Zhmoginov, and
  Chen}]{Sandler18}
Sandler, M.; Howard, A.; Zhu, M.; Zhmoginov, A.; and Chen, L.-C. 2018.
\newblock MobileNetV2: Inverted Residuals and Linear Bottlenecks.
\newblock \emph{arXiv preprint arXiv:1801.04381}.

\bibitem[{Shen et~al.(2021)Shen, Wang, Yin, Song, Luo, and Song}]{Shen21}
Shen, C.; Wang, X.; Yin, Y.; Song, J.; Luo, S.; and Song, M. 2021.
\newblock Progressive Network Grafting for Few-Shot Knowledge Distillation.
\newblock In \emph{AAAI}, volume~35, 2541--2549.

\bibitem[{Szegedy et~al.(2017)Szegedy, Ioffe, Vanhoucke, and Alemi}]{Szegedy17}
Szegedy, C.; Ioffe, S.; Vanhoucke, V.; and Alemi, A.~A. 2017.
\newblock Inception-v4, Inception-ResNet and the Impact of Residual Connections
  on Learning.
\newblock In \emph{AAAI}, 4278–4284.

\bibitem[{Tan and Le(2019)}]{tan19}
Tan, M.; and Le, Q. 2019.
\newblock {E}fficient{N}et: Rethinking Model Scaling for Convolutional Neural
  Networks.
\newblock In \emph{ICML}, volume~97, 6105--6114.

\bibitem[{Tan and Le(2021)}]{Tan21}
Tan, M.; and Le, Q.~V. 2021.
\newblock EfficientNetV2: Smaller Models and Faster Training.
\newblock \emph{arXiv preprint arXiv:2104.00298}.

\bibitem[{Tan, Pang, and Le(2020)}]{9156454}
Tan, M.; Pang, R.; and Le, Q.~V. 2020.
\newblock EfficientDet: Scalable and Efficient Object Detection.
\newblock In \emph{2020 IEEE/CVF Conference on Computer Vision and Pattern
  Recognition (CVPR)}, 10778--10787.

\bibitem[{Wang et~al.(2018)Wang, Zhao, Li, and Tan}]{wang18}
Wang, H.; Zhao, H.; Li, X.; and Tan, X. 2018.
\newblock Progressive Blockwise Knowledge Distillation for Neural Network
  Acceleration.
\newblock In \emph{IJCAI}, 2769--2775.

\bibitem[{Yim et~al.(2017)Yim, Joo, Bae, and Kim}]{Yim17}
Yim, J.; Joo, D.; Bae, J.; and Kim, J. 2017.
\newblock A Gift from Knowledge Distillation: Fast Optimization, Network
  Minimization and Transfer Learning.
\newblock In \emph{CVPR}, 7130--7138.

\bibitem[{Yu et~al.(2020)Yu, Jin, Liu, Bender, Kindermans, Tan, Huang, Song,
  Pang, and Le}]{yu20}
Yu, J.; Jin, P.; Liu, H.; Bender, G.; Kindermans, P.-J.; Tan, M.; Huang, T.;
  Song, X.; Pang, R.; and Le, Q. 2020.
\newblock BigNAS: Scaling up Neural Architecture Search with Big Single-Stage
  Models.
\newblock In \emph{ECCV}, 702--717.

\bibitem[{Yu, Mazaheri, and Jannesari(2021)}]{yu21}
Yu, S.; Mazaheri, A.; and Jannesari, A. 2021.
\newblock Auto Graph Encoder-Decoder for Neural Network Pruning.
\newblock In \emph{ICCV}, 6362--6372.

\bibitem[{Yun et~al.(2019)Yun, Han, Oh, Chun, Choe, and Yoo}]{Yun_2019_ICCV}
Yun, S.; Han, D.; Oh, S.~J.; Chun, S.; Choe, J.; and Yoo, Y. 2019.
\newblock CutMix: Regularization Strategy to Train Strong Classifiers With
  Localizable Features.
\newblock In \emph{ICCV}, 6022--6031.

\bibitem[{Zhang et~al.(2018)Zhang, Cisse, Dauphin, and
  Lopez-Paz}]{zhang2018mixup}
Zhang, H.; Cisse, M.; Dauphin, Y.~N.; and Lopez-Paz, D. 2018.
\newblock mixup: Beyond Empirical Risk Minimization.
\newblock In \emph{ICLR}.

\bibitem[{Zhong et~al.(2018)Zhong, Yan, Wu, Shao, and Liu}]{zhong18}
Zhong, Z.; Yan, J.; Wu, W.; Shao, J.; and Liu, C. 2018.
\newblock Practical Block-Wise Neural Network Architecture Generation.
\newblock In \emph{CVPR}, 2423--2432.

\bibitem[{Zhu et~al.(2021)Zhu, Wen, Du, Bian, Fan, Hu, and Ling}]{9573394}
Zhu, P.; Wen, L.; Du, D.; Bian, X.; Fan, H.; Hu, Q.; and Ling, H. 2021.
\newblock Detection and Tracking Meet Drones Challenge.
\newblock \emph{IEEE Transactions on Pattern Analysis and Machine
  Intelligence}, 1--1.

\bibitem[{Zhuang et~al.(2020)Zhuang, Tang, Ding, Tatikonda, Dvornek,
  Papademetris, and Duncan}]{Zhuang20}
Zhuang, J.; Tang, T.; Ding, Y.; Tatikonda, S.~C.; Dvornek, N.; Papademetris,
  X.; and Duncan, J. 2020.
\newblock AdaBelief Optimizer: Adapting Stepsizes by the Belief in Observed
  Gradients.
\newblock In \emph{NeurIPS}.

\end{thebibliography}

\appendix

\section{Remarks}

\subsection{Efficiency for Parallel Pretraining}
Recall that the pretraining step of Bespoke can be done in parallel, and its cost is $|\mathbf{A}|/G$ epochs.
LANA's pretraining step can also be done in parallel, but Bespoke requires much less cost than LANA for pretraining.
The reason for this comes from the number of targets (sub-networks) to train.
Suppose that an original neural network model has 100 layers.
LANA has 197 alternatives for each layer, so there are $100 \times 197$ mean-square-error losses.
On the other hand, Bespoke has 400 alternatives for the entire network, so there are only 400 mean-square-error losses.
This analysis implies that the number of training targets for LANA is an order of magnitude larger than that for Bespoke.

In addition, training many alternatives for a single layer requires high peak memory usage, which may hinder the efficiency of parallel training.
Bespoke, however, has only 400 alternatives over the entire network.
Because there were randomly selected, they are likely to be uniformly distributed.
Thus, roughly speaking, the peak memory usage of Bespoke with 400 alternatives is similar to that of LANA with four alternatives for every single layer.

In summary, Bespoke has a much less number of training targets than LANA.
Bespoke is also more efficient than LANA in peak memory usage, which is an essential factor for parallel training.
Those facts support the low cost of Bespoke for pretraining compared to LANA.

\begin{figure}[ht!]
	\centering
	\subfigure[Illustration of replacing a sub-network]{
		\includegraphics[scale =0.55] {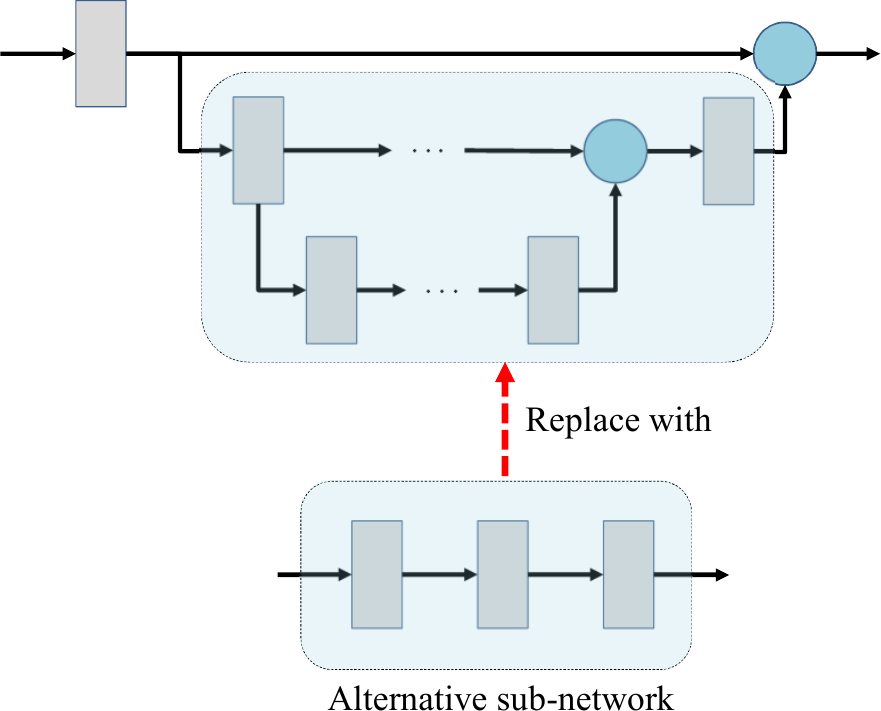}
		\label{fig:replace1}
	}
	\quad
	\subfigure[After the replacement]{
		\includegraphics[scale =0.55, angle=-90] {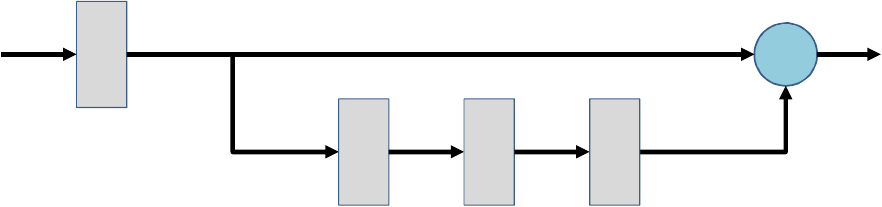}
		\label{fig:replace2}
	}
	\caption{Illustration of sub-network replacement.}
	\label{fig:replace}
\end{figure}

\subsection{Understanding Sub-Network Replacement}
Since this is the first work to replace a sub-network with another in a neural network model, one can be confused with the concept of the replacement.
For understanding, we provide an example of such a replacement in Figure~\ref{fig:replace1}.
Figure~\ref{fig:replace2} depicts the result of the replacement.
From this example, we want to clarify that a sub-network can be literally replaced with another.

\begin{figure*}
	\centering
	\subfigure[Two alternative blocks which have inconsistent in/out channels]{
		\includegraphics[scale =0.65] {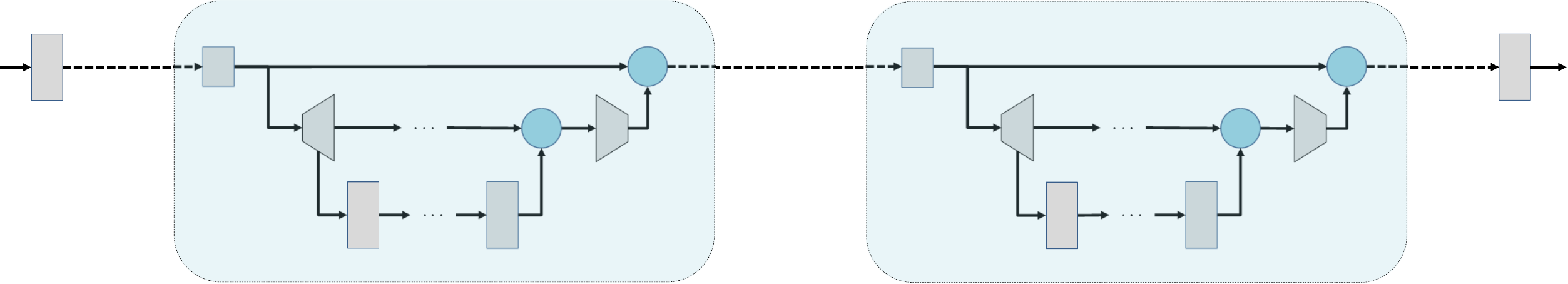}
		\label{fig:pc1u}
	}
	\quad
	\subfigure[Adding in/out masking layers for retraining]{
		\includegraphics[scale =0.65] {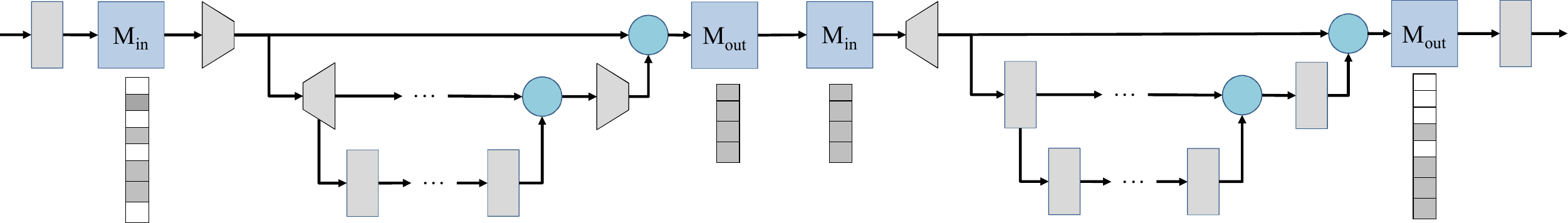}
		\label{fig:pc2u}
	}
	\quad
	\subfigure[Pruned sub-networks (after retraining)]{
		\includegraphics[scale =0.65] {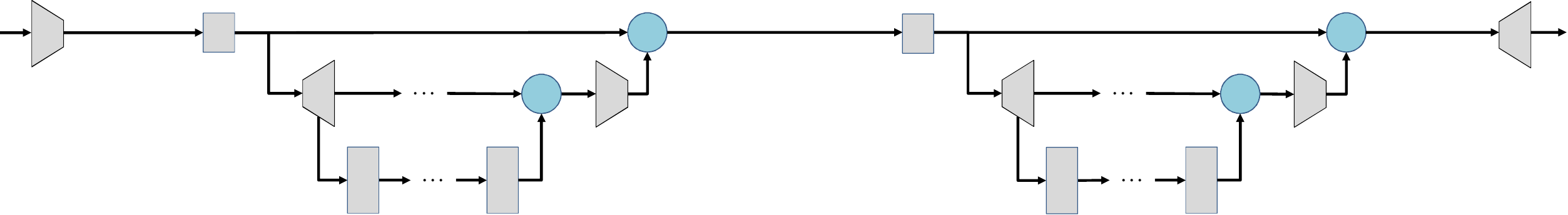}
		\label{fig:pc3u}
	}
	\caption{Illustration of adding in/out masking layers and pruning sub-networks.}
	\label{fig:masking}
\end{figure*}

\subsection{In/Out Masking Layers}
Figure~\ref{fig:masking} depicts the in/out masking layers of sub-networks.
Suppose that the two blocks identified by dotted boxes in Figure~\ref{fig:pc1u} are alternatives being added to the network.
The dotted arrow lines in this figure represent that the in/out channels of the blocks are inconsistent with the other parts of the network.
To make them consistent, we add masking layers as depicted Figure~\ref{fig:pc2u}.
Each masking layer has a masking vector for channel pruning.
After retraining the model having masking layers, we remove the masking layer and re-define the model based on the pruned channel information.
Then, the pruned model is given as the model in Figure~\ref{fig:pc3u}.

\section{Proofs}
\subsection{Proof for Lemma~1}
Since $l_{i}$ is the output layer of this sub-network and it has a single output, it is sufficient to show that the sub-network has a single input for proving this lemma.
Suppose that the sub-network has multiple inputs and there is a layer $l$ which is not included in the sub-network and provides an input data to another $l'$ in the sub-network.
It implies that $l'$ was popped from $S$ before $l_{i}$ is popped in Algorithm~1.
However, since $l$ is not included in $P_{i}$, $l$ is never added to $S$.
This implies that $l$ is never popped from $S$ and $\delta[l']$ is not increased via $l$, which means that $\delta[l'] < |\mathbf{N}_{\text{in}}(l')|$.
Thus, $l'$ cannot be popped from $S$, which contradicts the assumption that $l'$ is a layer in the sub-network.
By contradiction, the sub-network must have a single input, so the lemma is proved.

\subsection{Proof for Theorem~1}
By Lemma~1, every sub-network found by Algorithm~1 has a single input and a single output.
Then, the remaining thing for proving Theorem~1 is that all sub-networks having a single input and a single output can be found by Algorithm~1.

Let us assume that we have a sub-network having a single input and a single output that cannot be found by Algorithm~1.
Let us denote the starting layer of the sub-network by $l$ and its output layer by $l'$.
Since Algorithm~1 can start from any layer, we can easily make it start from $l$.
Then, it implies that $l'$ cannot be added into $S$ in Algorithm~1 even if it starts from $l$ under the assumption.
In this situation, there are two possible cases.
One case is that $l$ and $l'$ are disconnected from each other, but it is not our case because they are included in the same sub-network with a single input and a single output.
The other is that $\delta[l'] < |\mathbf{N}_{\text{in}}(l')|$.
In other words, there is an inbound layer $\hat{l}$ of $l'$, which is never added into $S$.
The only possible reason for this is that there is another inbound layer of $\hat{l}$ which is never added into $S$.
In this way, we can follow layers in the inbound direction to find a certain initial layer.
Note that the initial layer cannot be $l$ because $l$ is popped from $S$ according to Algorithm~1.
Then, it should be a leaf layer, which is an input layer of the entire neural network $\mathcal{T}$.
However, since we assumed that the sub-network has a single input, such a leaf layer does not exist.
This analysis demonstrates that such a sub-network cannot exist.
By contracting the assumption, all sub-networks having a single input and a single output can be found by Algorithm~1.
Therefore, Theorem~1 is proved.

\begin{table}[th!]
	\centering
	\caption{Parameter Sensitivity Test for $\lambda$ (CIFAR-100, Bespoke-B2)}
	\begin{tabular}[c]{l | c | c}
		\hline
		
		Ratio		&	Top-1 Acc. (\%)	&	CPU Latency (ms)	\\
		\hline
		\hline
		1.0				&	\bf{85.45}		&	31.82		\\
		0.5				&	85.35		&	28.15		\\		
		0.0				&	85.21		&	\bf{21.73} 	\\		
		\hline
	\end{tabular}
	\centering
	\label{table:lambda}
\end{table}

\section{More Empirical Results}
\subsection{Parameter Testing with $\lambda$}

Equation~5 of the manuscript includes a real-valued parameter $\lambda \geq 0$.
We conducted a sensitivity test for this, and the result is shown in Table~\ref{table:lambda}.
As $\lambda$ gets smaller, the accuracy of the result model gets lower and the inference speed of it is faster.
One can think that the result accuracy is reasonable and useful when $\lambda=0$.
Nevertheless, because such a reasonable accuracy is not always guaranteed,
we usually set $\lambda$ to be from $0.5$ to $1.0$.

\begin{table}[th!]
	\centering
	\caption{Parameter Sensitivity Test for $r$ (CIFAR-100, Bespoke-B2)}
	\begin{tabular}[c]{l | c | c}
		\hline
		Ratio		&	Top-1 Acc. (\%)	&	CPU Latency (ms)	\\
		\hline
		\hline
		0.1		&	84.42			&	40.93 	\\		
		0.3		&	\bf{85.45}		&	\bf{31.82}		\\
		\hline
	\end{tabular}
	\centering
	\label{table:lambda}
\end{table}

\subsection{Parameter Testing with $r$}
Recall $r$ is the ratio parameter to control the maximum size of sampled sub-networks.
We also conducted parameter testing with $r$.
We did not conduct a sufficient number of variations for $r$, but the difference between the two slots is still meaningful.

When $r$ is 0.1, the number of layers in a sub-network is likely to be small.
Bespoke with $r=0.1$ requires less memory usage in preprocessing, but its optimization performance gets poor.
This may be supporting evidence for why we need to consider a sub-network (block), not a layer.

\begin{table}[t]
	\centering
	\small
	\caption{TensorRT Latency (ImageNet)}
	\begin{tabular}[c]{l | c | c}
		\hline
		Methods		&	Top-1 Acc. (\%)	&	Latency (ms)		 \\
		\hline
		\hline
		EfficientNetB2						&		79.98			&	11.52		\\
		EfficientNetB0						&		77.19			&	6.71		\\		
		\bf{Bespoke-EB2}					&		\bf{78.08}		& 	\bf{7.10}	\\								
		\hline
		MobileNetV3-Large					&		75.68			&	4.07		\\				
		MobileNetV3-Small					&		68.21			&	2.96	\\
		CURL								&		71.21			&	6.08		\\										
		\bf{Bespoke-EB0}					&		\bf{72.75}		&	\bf{6.13}		\\
		\hline
	\end{tabular}
	\centering
	\label{table:trt}
\end{table}

\subsection{Edge-Level Inference}
We conducted an experiment to see the latency of models found by Bespoke in an edge device named NVIDIA Jetson Xaiver with TensorRT.
The result is shown in Table~\ref{table:trt}.
Note that the models tested in this experiment were founded via ONNX-based GPU latency.
The result says that Bespoke-EB2 has comparable inference time to EfficientNet-B0 with a higher accuracy.

The inference speed of Bespoke-EB0 is slower than CURL and MobileNetV3 models.
We need to do further study why it happens and how to resolve this issue.

\subsection{Object Detection Task}
To demonstrate the generality of Bespoke, we applied Bespoke to an object detection task.
For this experiment, we use a small object detection dataset named VisDrone \cite{9573394}.
VisDrone includes images taken by drones.
The number of the images of VisDrone is 10,209 images, and there are million-level bounding boxes based on ten categories.
The used object detection model is EfficientDet \cite{9156454}.
We trained EfficientDet for VisDrone based on the implementation provided in here\footnote{https://github.com/google/automl}.
Note that Bespoke is applied to accelerating EfficientDet by replacing the backbone of it with what is produced by Bespoke.

The results are described in Table~\ref{table:trt-ob}.
Bespoke-D2 (Bespoke-D0) is the model obtained by applying Bespoke to EfficientDet-D2 (Bespoke-D0).
While the latency of Bespoke-D2 is similar to that of Efficient-D0, Bespoke-D2 is somewhat behind in terms of average precision.
There are several possible reasons.
First, we did not use any hyperparameter optimization technique for training EfficientDet for VisDrone.
Thus, the base accuracy of EfficientDet-D2 is lower than expected.
Note that the gap between the average precisions of EfficientDet-D2 and EfficientDet-D0 is about 10 for the COCO dataset \cite{lin2014microsoft},
while it is only 1.5 in this experiment.
We think that if we find a more appropriate hyperparameter set for EfficientDet-D2, Bespoke-D2 will be more effective than EfficientDet-D0.
Nevertheless, even if we change only the backbone of EfficientDet, the latency gain must be significant with respect to the loss of accuracy.

\begin{table}[t]
	\centering
	\small
	\caption{Object Detection Result}
	\begin{tabular}[c]{l | c | c}
		\hline
		Methods		&	AP$_{50-95}$	&	Latency (ms)		 \\
		\hline
		\hline
		EfficientDet-D2						&		19.5			&	809.41		\\
		\bf{Bespoke-D2}						&		16.4			& 	\bf{548.00}	\\								
		\hline
		EfficientDet-D0						&		18.0			&	546.70		\\				
		\bf{Bespoke-D0}						&		12.4			&	\bf{334.73}	\\
		\hline
	\end{tabular}
	\centering
	\label{table:trt-ob}
\end{table}

\subsection{Visualization}
\begin{figure*}
	\centering
	\subfigure[Bespoke-B2 for CPU]{
		\includegraphics[scale =0.75] {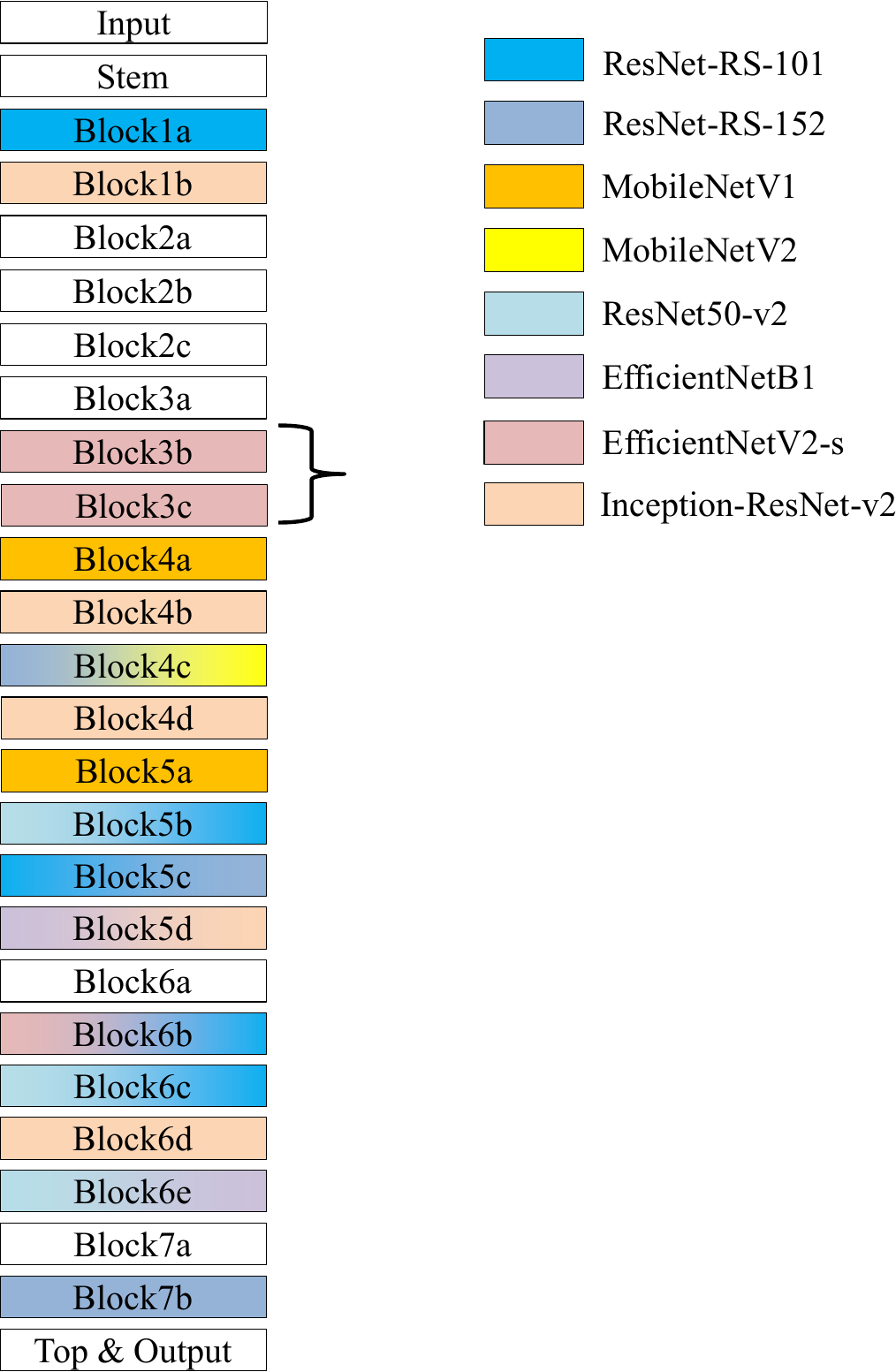}
		\label{fig:cifar100_cpu}
	}
	\qquad
	\subfigure[Bespoke-B2 for GPU]{
		\includegraphics[scale =0.75] {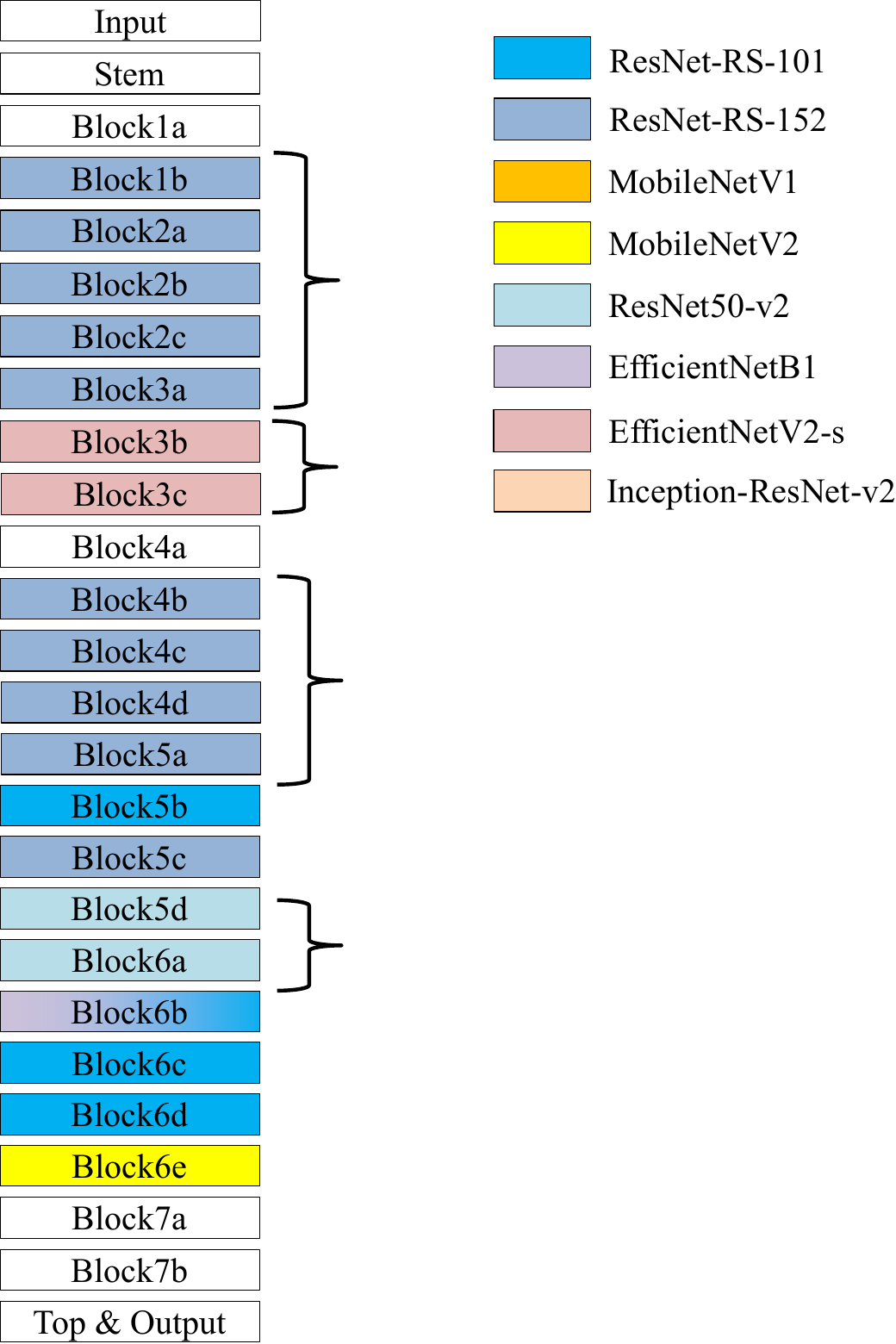}
		\label{fig:cifar100_gpu}
	}
	\caption{Model visualization of Bespoke. If a rectangular block in EfficientNet-B2 is fully or partially changed by Bespoke, the block is colored with the color corresponding to a pretrained model where the replacement comes.}
	\label{fig:models}
\end{figure*}

We provide visualization for models produced by Bespoke.
Figure~\ref{fig:models} depicts Bespoke-B2 models for CIFAR-100, which were optimized for CPU or GPU.
Rectangular blocks in this figure are the blocks of EfficientNet-B2.
If a rectangular block is partially or fully replaced by Bespoke, it is colored with the color corresponding to a pretrained model where the replacement comes.
Note that a block can be partially replaced multiple times because every block has many layers.
In this case, it is colored with gradation color.
There are several brackets each of which means that multiple blocks in EfficientNet-B2 are replaced together with a single alternative block by Bespoke. 

While Figure~\ref{fig:cifar100_cpu} shows Bespoke-B2 optimized for CPU inference, Figure~\ref{fig:cifar100_gpu} shows it for GPU inference.
The replacement results of the two cases are quite different, and this fact yields the reason why we need to consider environments when searching models.
Note that for the GPU case, there are many consecutive blocks replaced together with efficient replacements.
Replacing such blocks with a single block is likely to make the result model fast, because it can work like layer skipping.

\end{document}